\documentclass{article}

\PassOptionsToPackage{numbers, square, sort&compress}{natbib}
\usepackage{iclr2020_conference,times}

\usepackage[utf8]{inputenc} % allow utf-8 input
\usepackage[T1]{fontenc}    % use 8-bit T1 fonts
\usepackage{hyperref}       % hyperlinks
\usepackage{url}            % simple URL typesetting
\usepackage{booktabs}       % professional-quality tables
\usepackage{amsfonts}       % blackboard math symbols
\usepackage{nicefrac}       % compact symbols for 1/2, etc.
\usepackage{microtype}      % microtypography
\usepackage{amsmath}
\usepackage{amsthm}
\usepackage{multirow}
\usepackage{caption}
\usepackage{graphicx}
\usepackage{siunitx}
\usepackage{mathtools}
\usepackage{xcolor}
\usepackage{wrapfig}
\usepackage{xcolor}
\usepackage{amssymb}
\usepackage[symbol,para]{footmisc}

\renewcommand{\vec}[1]{\mathbf{#1}}
\newcommand{\set}[1]{#1}
\newcommand{\method}{HOF}
\newcommand{\img}{I}
\newcommand{\Enc}{g_\phi}
\newcommand{\DecNoParam}{f_\theta}
\newcommand{\FuncName}{mapping}

\newcommand{\Dec}[1]{
    \ifthenelse{\equal{#1}{}}{f_\theta}{f_{\theta_{#1}}}
}
\newcommand{\R}{\mathbb{R}}
\newcommand{\Obj}{O}

\newcommand{\website}{https://saic-ny.github.io/hof}

\title{Higher-Order Function Networks for Learning Composable 3D Object Representations}

\makeatletter
\renewcommand*{\@fnsymbol}[1]{\ensuremath{\ifcase#1\or 1 \or 2 \or 3 \or \dagger \or \ddagger \else\@ctrerr\fi}}
\makeatother

\author{Eric Mitchell\thanks{Stanford University}~~\thanks{Samsung AI Center - New York} \And Selim Engin\thanks{University of Minnesota}~~\thanks{Work performed while an intern at Samsung AI Center - New York.} \And Volkan Isler\footnotemark[2] \And Daniel D Lee\footnotemark[2]
}

\iclrfinalcopy
\begin{document}

\maketitle

\begin{abstract}
We present a new approach to 3D object representation where a neural network encodes the geometry of an object directly into the weights and biases of a second `mapping' network. This mapping network can be used to reconstruct an object by applying its encoded transformation to points randomly sampled from a simple geometric space, such as the unit sphere. We study the effectiveness of our method through various experiments on subsets of the ShapeNet dataset. We find that the proposed approach can reconstruct encoded objects with accuracy equal to or exceeding state-of-the-art methods with orders of magnitude fewer parameters. Our smallest mapping network has only about 7000 parameters and shows reconstruction quality on par with state-of-the-art object decoder architectures with millions of parameters. Further experiments on feature mixing through the composition of learned functions show that the encoding captures a meaningful subspace of objects. \footnote[5]{See \href{\website}{\texttt{\website}} for code and additional information. Correspondence to: Eric Mitchell <eric.mitchell@cs.stanford.edu>.}
\end{abstract}

%%%%%%%%%%%%%%%%%%%%%%%%%%%%%%%%%%%%%%%%%%%%%%%%%%%%%%%%%%%%%%%%%%%%%%%%%%%%%%%%%%%%%%%%
%%%%%%%%%%%%%%%%%%%%%%%%%%%%%%%%%%%%%%%%%%%%%%%%%%%%%%%%%%%%%%%%%%%%%%%%%%%%%%%%%%%%%%%%
\section{Introduction}
%%%%%%%%%%%%%%%%%%%%%%%%%%%%%%%%%%%%%%%%%%%%%%%%%%%%%%%%%%%%%%%%%%%%%%%%%%%%%%%%%%%%%%%%
%%%%%%%%%%%%%%%%%%%%%%%%%%%%%%%%%%%%%%%%%%%%%%%%%%%%%%%%%%%%%%%%%%%%%%%%%%%%%%%%%%%%%%%%

This paper is primarily concerned with the problem of learning compact 3D object representations and estimating them from images. If we consider an object to be a continuous surface in $\mathbb{R}^3$, it is not straightforward to directly represent this infinite set of points in memory. In working around this problem, many learning-based approaches to 3D object representation suffer from problems related to memory usage, computational burden, or sampling efficiency. Nonetheless, neural networks with tens of millions of parameters have proven effective tools for learning expressive representations of geometric data. In this work, we show that object geometries can be encoded into neural networks with thousands, rather than millions, of parameters with little or no loss in reconstruction quality.

To this end, we propose an object representation that encodes an object as a function that maps points from a canonical space, such as the unit sphere, to the set of points defining the object. In this work, the function is approximated with a small multilayer perceptron. The parameters of this function are estimated by a `higher order' encoder network, thus motivating the name for our method: \textit{Higher-Order Function networks (\method{})}. This procedure is shown in Figure 1. There are two key ideas that distinguish HOF from prior work in 3D object representation learning: fast-weights object encoding and interpolation through function composition.

\textit{(1) Fast-weights object encoding:} `Fast-weights' in this context generally refers to methods that use network weights and biases that are not fixed; at least some of these parameters are estimated on a per-sample basis. Our fast-weights approach stands in contrast to existing methods which encode objects as vector-valued inputs to a decoder network with fixed weights. Empirically, we find that our approach enables a dramatic reduction (two orders of magnitude) in the size of the \FuncName{} network compared to the decoder networks employed by other methods.

\textit{(2) Interpolation through function composition:} Our functional formulation allows for interpolation between inputs by composing the roots of our reconstruction functions. We demonstrate that the functional representation learned by HOF provides a rich latent space in which we can `interpolate' between objects, producing new, coherent objects sharing properties of the `parent' objects.
% for interpolation differs from existing
% work, which primarily uses linear interpolation of latent feature vectors (the
% inputs to a decoder function) to interpolate between objects

In order to position HOF among other methods for 3D reconstruction, we first define a taxonomy of existing work and show that HOF provides a generalization of current best-performing methods. Afterwards, we demonstrate the effectiveness of \method{} on the task of 3D reconstruction from an RGB image using a subset of the ShapeNet dataset~\citep{chang2015shapenet}. The results, reported in Tables~\ref{tab:mainresults} and~\ref{tab:what3d-results} and Figure~\ref{fig:recon-compare}, show state-of-the-art reconstruction quality using orders of magnitude fewer parameters than other methods.

\begin{figure}%[ht]
  \centering
  \includegraphics[width=0.99\textwidth]{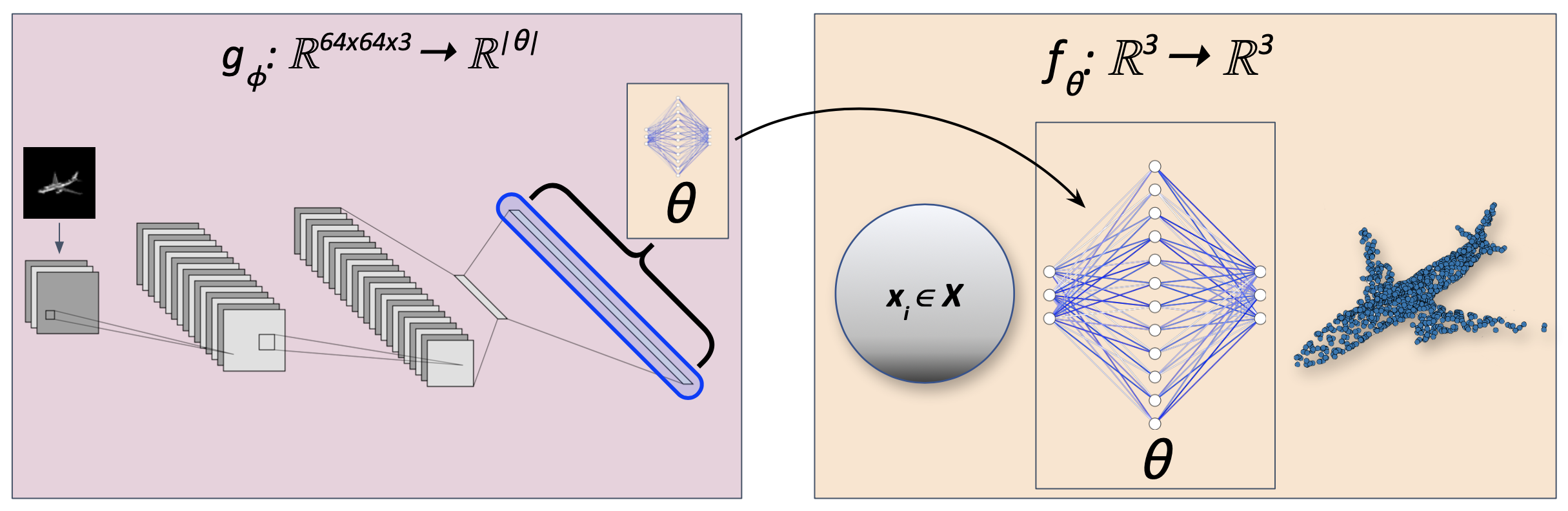}
  \includegraphics[width=0.92\textwidth]{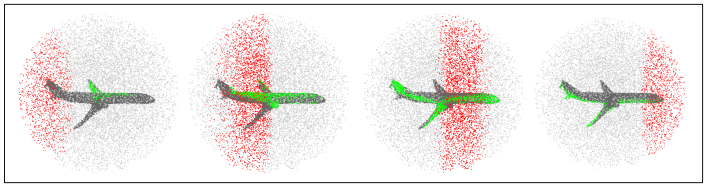}
  \caption{\textbf{Top}: Overview of HOF. The encoder network $\Enc$ encodes the geometry of the object pictured in each input image directly into the parameters of the mapping function $\DecNoParam$, which produces a reconstruction as a transformation of a canonical object (here, the unit sphere). \textbf{Bottom}: We visualize the transformation $f_\theta$ by showing various subsets of the inputs $X$ and their corresponding mapped locations in red and green, respectively. In each frame, light gray shows the rest of $X$ and dark gray shows the rest of the reconstructed object.} \label{fig:overview_fig}
\end{figure}

% We also evaluate \method{} for a motion planning application where the objective is to navigate around an object whose image from the initial location is given. In order to compare object reconstruction methods, we generate a 3D model using this input image and plan a shortest path around the model with the A$^*$ algorithm.  Experiments indicate that using reconstructions produced by \method{} for path planning yields shorter and safer paths.

%%%%%%%%%%%%%%%%%%%%%%%%%%%%%%%%%%%%%%%%%%%%%%%%%%%%%%%%%%%%%%%%%%%%%%%%%%%%%%%%%%%%%%%%
%%%%%%%%%%%%%%%%%%%%%%%%%%%%%%%%%%%%%%%%%%%%%%%%%%%%%%%%%%%%%%%%%%%%%%%%%%%%%%%%%%%%%%%%
\section{Related Work}
%%%%%%%%%%%%%%%%%%%%%%%%%%%%%%%%%%%%%%%%%%%%%%%%%%%%%%%%%%%%%%%%%%%%%%%%%%%%%%%%%%%%%%%%
%%%%%%%%%%%%%%%%%%%%%%%%%%%%%%%%%%%%%%%%%%%%%%%%%%%%%%%%%%%%%%%%%%%%%%%%%%%%%%%%%%%%%%%%

The selection of object representation is a crucial design choice for methods addressing 3D reconstruction. Voxel-based approaches~\citep{choy2016r2n2,hane2017hierarchical} typically use a uniform discretization of $\mathbb{R}^3$ in order to extend highly successful convolutional neural network (CNN) based approaches to three dimensions. However, the inherent sparsity of surfaces in 3D space make voxelization inefficient in terms of both memory and computation time. Partition-based approaches such as octrees~\citep{tatar2017octree,RieglerUG16} address the space efficiency shortcomings of voxelization, but they are tedious to implement and more computationally demanding to query. 
Graph-based models such as meshes~\citep{wang2018pixel2mesh, gkioxari2019meshrcnn, smith2019geometrics,hanocka2019meshcnn} provide a compact representation for capturing topology and surface level information, however their irregular structure makes them harder to learn.
Point set representations, discrete (and typically finite) subsets of the continuous geometric object, have also gained popularity due to the fact that they retain the simplicity of voxel based methods while eliminating their storage and computational burden~\citep{qi2016point,fan2017point,qi2017pointnetplusplus,yang2018foldingnet,park2019deepsdf}. The PointNet architecture \citep{qi2016point,qi2017pointnetplusplus} was an architectural milestone that made manipulating point sets with deep learning methods a competitive alternative to earlier approaches; however, PointNet is concerned with \textit{processing}, rather than \textit{generating}, point clouds. Further, while point clouds are more flexible than voxels in terms of information density, it is still not obvious how to adapt them to the task of producing arbitrary- or varied-resolution predictions. Independently regressing each point in the point set requires additional parameters for each additional point \citep{fan2017point,achlioptas2017learning}, which is an undesirable property if the goal is high-resolution point clouds.

Many current approaches to representation and reconstruction follow an encoder-decoder paradigm, where the encoder and decoder both have learned weights that are fixed at the end of training. An image or set of 3D points is encoded as a latent vector `codeword' either with a learned encoder as in~\cite{yang2018foldingnet,lin2018learning,yan2016perspective} or by direct optimization of the latent vector itself with respect to a reconstruction-based objective function as in~\cite{park2019deepsdf}. Afterwards, the latent code is decoded by a learned decoder into a reconstruction of the desired object by one of two methods, which we call \textit{direct decoding} and \textit{contextual mapping}. Direct decoding methods directly map the latent code into a fixed set of points~\citep{fan2017point,choy2016r2n2,machalkiewicz2019deep,lin2018learning}; contextual mapping methods map the latent code into a function that can be sampled or otherwise manipulated to acquire a reconstruction~\citep{park2019deepsdf,yang2018foldingnet,mescheder2018occupancy,machalkiewicz2019deep}. Direct decoding methods generally suffer from the limitation that their predictions are of fixed resolution; they cannot be sampled more or less precisely. With contextual mapping methods, it is possible in principle to sample the object to arbitrarily high resolution with the correct decoder function. However, sampling can provide a significant computational burden for some contextual mapping approaches as those proposed by \cite{park2019deepsdf} and \cite{machalkiewicz2019deep}. Another hurdle is the need for post-processing such as applying the Marching Cubes algorithm developed by \cite{lorensen1987marching}. We call contextual mapping approaches that encode context by concatenating a duplicate of a latent context vector with each input \textit{latent vector concatenation (LVC)} methods. In particular, we compare with LVC architectures used in FoldingNet~\citep{yang2018foldingnet} and DeepSDF~\citep{park2019deepsdf}.

HOF is a contextual mapping method that distinguishes itself from other methods within this class through its approach to representing the mapping function: HOF uses one neural network to estimate the weights of another. Conceptually related methods have been previously studied under nomenclature such as the `fast-weight' paradigm~\citep{schmiudhuber1992learning,debrabandere16dynamic,Klein2015,Riegler2015} and more recently `hypernetworks'~\citep{ha2016hyper}. However, the work by~\cite{schmiudhuber1992learning} deals with encoding memories in sequence learning tasks. \cite{ha2016hyper} suggest that estimating weights of one network with another might lead to improvements in parameter-efficiency. However, this work does not leverage the key insight of using network parameters that are estimated \textit{per sample} in vision tasks.

\label{sec:taxonomy}

\begin{figure}
  \centering
  \includegraphics[width=0.8\textwidth]{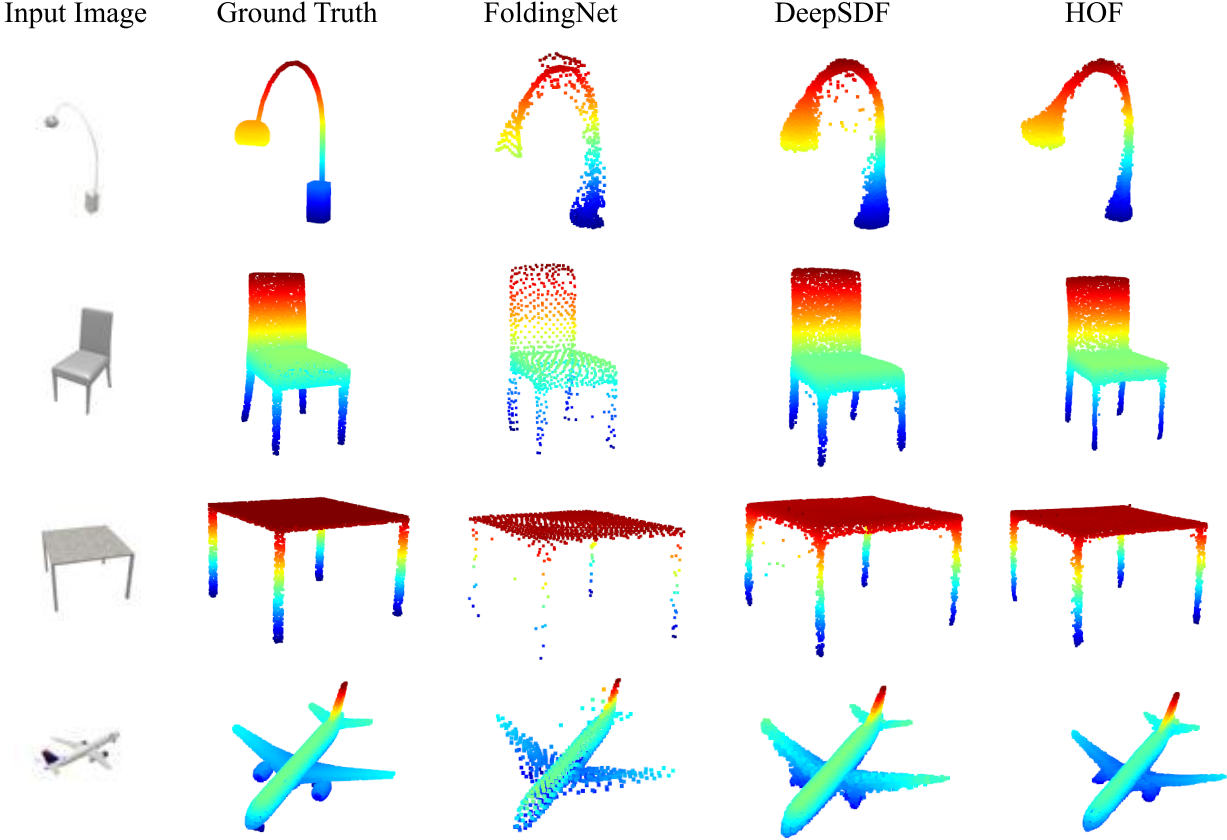}
  \caption{From left to right: Input RGB image, ground truth point cloud, reconstruction from FoldingNet~\citep{yang2018foldingnet}, reconstruction from DeepSDF~\citep{park2019deepsdf}, and our method.}
  \label{fig:recon-compare}
\end{figure}

%%%%%%%%%%%%%%%%%%%%%%%%%%%%%%%%%%%%%%%%%%%%%%%%%%%%%%%%%%%%%%%%%%%%%%%%%%%%%%%%%%%%%%%%
%%%%%%%%%%%%%%%%%%%%%%%%%%%%%%%%%%%%%%%%%%%%%%%%%%%%%%%%%%%%%%%%%%%%%%%%%%%%%%%%%%%%%%%%
\section{Higher-Order Function Networks}

%%%%%%%%%%%%%%%%%%%%%%%%%%%%%%%%%%%%%%%%%%%%%%%%%%%%%%%%%%%%%%%%%%%%%%%%%%%%%%%%%%%%%%%%
%%%%%%%%%%%%%%%%%%%%%%%%%%%%%%%%%%%%%%%%%%%%%%%%%%%%%%%%%%%%%%%%%%%%%%%%%%%%%%%%%%%%%%%%

% The object representation that we propose is characterized by a mapping function $\Dec{}$ that maps a canonical set $\set{X}$, such as the interior of the unit sphere, to the set of 3D points that define the surface or interior of an object $\Obj$. The mapping function is represented by a multilayer perceptron (MLP). 

HOF is motivated by the independent observations by both \cite{yang2018foldingnet} and \cite{park2019deepsdf} that LVC methods do not perform competitively when the context vector is injected by simply concatenating it with each input. In both works, the LVC methods proposed required architectural workarounds to produce sufficient performance on reconstruction tasks, including injecting the latent code multiple times at various layers in the network. HOF does not suffer from these shortcomings due to its richer context encoding (the entire mapping network encodes context) in comparison with LVC. We compare the HOF and LVC regimes more precisely in Section~\ref{sec:comparing}. Quantitative comparisons of HOF with existing methods can be found in Table~\ref{tab:mainresults}.

% Instead of estimating a latent context vector for each input image, a HOF network directly estimates the entire mapping function $\Dec{}$ for an input image. This approach is not limited to images; using an encoder architecture like PointNet~\citep{qi2016point} enables encoding of partial point clouds, for example. In our formulation, in contrast to the direct decoding and contextual mapping methods mentioned previously, the output of our \FuncName{} network is modulated by simply changing the connection strengths within the network.  Next, we formalize HOF for object reconstruction.

\subsection{A Fast-Weights Approach to 3D Object Representation and Reconstruction}

We consider the task of reconstructing an object point cloud $\Obj$ from an image. We start by training a neural network $\Enc$ with parameters $\phi$ (Figure~\ref{fig:overview_fig}, top-left) to output the parameters $\theta$ of a mapping function $\DecNoParam$, which reconstructs the object when applied to a set of points $X$ sampled uniformly from a canonical set such as the unit sphere (Figure~\ref{fig:overview_fig}, top-right). We note that the number of samples in $X$ can be increased or decreased to produce higher or lower resolution reconstructions without changing the network architecture or retraining, in contrast with direct decoding methods and some contextual mapping methods which use fixed, non-random samples from $X$ \citep{yang2018foldingnet}. The input to $\Enc$ is an RGB image $I$; our implementation takes $64 \times 64 \times 3$ RGB images as input, but our method is general to any input representation for which a corresponding differentiable encoder network can be constructed to estimate $\theta$ (e.g. PointNet \citep{qi2016point} for point cloud completion). Given $I$, we compute the parameters of the \FuncName{} network $\theta_I$ as
\begin{align}
    \theta_I = \Enc(\img)
\end{align}
That is, the encoder $\Enc : \R^{3\times 64 \times 64} \rightarrow \R^d$ directly regresses the $d$-dimensional parameters $\theta_I$ of the \FuncName{} network $\Dec{I}:\R^c \rightarrow \R^3$, which maps $c$-dimensional points in the canonical space $\set{X}$ to points in the reconstruction $\hat{\Obj}$ (see Figure~\ref{fig:overview_fig}). We then transform our canonical space $\set{X}$ with $\Dec{I}$ in the same manner as other contextual mapping methods: 
\begin{align}
    \hat{\Obj} = \{f_{\theta_I}(\mathbf{x_i}): \mathbf{x_i} \in X\}\label{eq:first-power}
\end{align}

% As a contextual mapping method, HOF allows for higher- or lower-resolution sampling of the reconstructed object in an on-line fashion by continually sampling points from $\set{X}$ as desired. In this work, $\set{X}$ corresponds to points within the unit sphere in $\R^3$, that is: $\set{X} = \{\vec{x} : ||\vec{x}||_2 \le 1\}$. However, we have found that the performance of \method{} is not very sensitive to the choice of $\set{X}$; sampling from the surface of the unit sphere in $\R^3$ or the interior of a 512-dimensional unit hypercube yielded similar-quality reconstructions.

During training, we sample an image $I$ and the corresponding ground truth point cloud model $\Obj$, where $\Obj$ contains 10,000 points sampled from the surface of the true object. We then obtain the \FuncName{} $\Dec{I}=\Enc(I)$ and produce an estimated reconstruction of $\Obj$ as in Equation~\ref{eq:first-power}. In our training, we only compute $\Dec{I}(\vec{x})$ for a sample of 1000 points in $\set{X}$. However, we find that sampling many more points (10-100$\times$ as many) at test time still yields high-quality reconstructions. This sample is drawn from a uniform distribution over the set $X$. We then compute a loss for the prediction $\hat{\Obj}$ using a differentiable set similarity metric such as Chamfer distance or Earth Mover's Distance. We focus on the Chamfer distance as both a training objective and metric for assessing reconstruction quality. The asymmetric Chamfer distance $CD(\set{X}, \set{Y})$ is often used for quantifying the similarity of two point sets $\set{X}$ and $\set{Y}$ and is given as

\begin{equation}
\label{eq:chamfer}
    CD(\set{X}, \set{Y}) = \frac{1}{|\set{X}|}\sum_{\vec{x}_i\in\set{X}}\min_{\vec{y}_i\in\set{Y}}||\vec{x}_i - \vec{y}_i||^2_2
\end{equation}

The Chamfer distance is defined even if sets $X$ and $Y$ have different cardinality. We train $\Enc$ to minimize the \textit{symmetric} objective function $\ell(\hat{\Obj}, \Obj) = CD(\hat{\Obj}, \Obj) + CD(\Obj, \hat{\Obj})$ as in \cite{fan2017point}.

\subsection{Comparison with LVC Methods}
\label{sec:comparing}

We compare our \FuncName{} approach with LVC architectures such as DeepSDF~\citep{park2019deepsdf} and FoldingNet~\citep{yang2018foldingnet}. These architectures control the output of the decoder through the concatenation of a latent `codeword' vector $\vec{z}$ with each input $\vec{x_i} \in \set{X}$. The codeword is estimated by an encoder $g_{\phi_\text{LVC}}$ for each image. We consider the case in which the latent vector is only concatenated with inputs in the first layer of the decoder network $\Dec{}$, which we assume to be an MLP. We are interested in analyzing the manner in which the network output with respect to $\vec{x_i}$ may be modulated by varying $\vec{z}$.

If the vector $\vec{a_i}$ contains the pre-activations of the first layer of $\Dec{}$ given an input point $\vec{x_i}$, we have
\begin{equation*}
    \vec{a_i} = W^\vec{x}\vec{x_i} + W^\vec{z}\vec{z} + \vec{b}
\end{equation*}

where $W^\vec{x}$, $W^\vec{z}$, and $\vec{b}$ are fixed parameters of the decoder, and only $\vec{z}$ is a function of $I$. If we absorb the parameters $W^\vec{z}$ and $\vec{b}$ into the encoder parameters $\phi_\text{LVC}$ (as $W^\vec{z}$ and $\vec{b}$ are fixed for all $\vec{x_i}$), we can define a new, equivalent latent representation $\vec{b}^*=W^\vec{z}\vec{z} + \vec{b}=W^\vec{z}g_{\phi_\text{LVC}}(I) + \vec{b}$ and a new encoder function $h$ with parameters $\phi_\text{LVC} \cup \{W^\vec{z}, \vec{b}\}$ such that $h(I)=\vec{b}^*$. This gives
\begin{equation*}
    \vec{a_i} = W^\vec{x}\vec{x_i} + h(I)
\end{equation*}

Thus the LVC approach is equivalent to estimating a \textit{fixed subset} of the parameters $\theta$ of the decoder $\Dec{}$ on a per-sample basis (the first layer bias). From this perspective, \method{} is an intuitive generalization: rather than estimating just the first layer bias, we allow our encoder to modulate all of the parameters in the decoder $\Dec{}$ on a per-sample basis.

Having demonstrated HOF as a generalization of existing contextual mapping methods,
in the next section, we present a novel application of contextual mapping that leverages the compositionality of the estimated mapping functions to aggregate features of multiple objects or multiple viewpoints of the same object.

\subsection{Extending Contextual Mapping Methods: Feature Aggregation through Function Composition}

An advantageous property of methods that use a latent codeword is that they have been empirically shown to learn a meaningful space of object geometries, in which interpolating between object encodings gives new, coherent object encodings \citep{yang2018foldingnet,park2019deepsdf,groueix2018papier,fan2017point}. HOF, on the other hand, does not obviously share this property: interpolating between the mapping function parameters estimated for two different objects need not yield a new, coherent object as the prior work has shown that the solution space of `good' neural networks is highly non-convex \citep{li2018visualizing}. We demonstrate empirically in Figure~\ref{fig:interp-comparison} that naively interpolating between reconstruction function in the HOF regime does indeed produce meaningless blobs. However, with a small modification to the HOF formulation in Equation~\ref{eq:first-power}, we can in fact learn a rich space of \textit{functions} in which we can interpolate between objects through function composition. 

We extend the formulation in Equation \ref{eq:first-power} to one where an object is represented as the $k$-th power of the \FuncName{} $\Dec{I}$:
\begin{align}
    \hat{\Obj} = \{\Dec{I}^k(\mathbf{x}): \mathbf{x} \in X\}\label{eq:kth-power}
\end{align}
where $f^k$ is defined as the composition of $f$ with itself $(k-1)$ times: $f^k(\vec{x}) = f(f^{(k-1)}(\vec{x}))$ where $f^0(\vec{x})\triangleq\vec{x}$. We call a \FuncName{} $\Dec{I}$ whose $k$-th power reconstructs the object $\Obj$ in image $I$ the $k$-\FuncName{} for $\Obj$.

This modification to Equation~\ref{eq:first-power} adds an additional constraint to the \FuncName: the domain and codomain must be the same. However, evaluating powers of $f$ leverages the power of weight sharing in neural network architectures; for an MLP \FuncName{} architecture with $l$ layers (excluding the input layer), evaluating its $k$-th power is equivalent to an MLP with $l\times k$ layers with shared weights. This formulation also has connections to  earlier work on continuous attractor networks as a model for encoding memories in the brain as $k$ becomes large~\citep{seung1998continuous}.

In Section \ref{sec:composition}, we conduct experiments in a setting in which we have acquired RGB images $I$ and $J$ of two objects, $\Obj_I$ and $\Obj_J$, respectively. Applying our encoder to these images, we obtain $k$-\FuncName s $\Dec{I}$ and $\Dec{J}$, which have parameters $\theta_{I} = \Enc(I)$ and $\theta_{J} = \Enc(J)$, respectively. We hypothesize that we can combine the information contained in each mapping function $\Dec{i}$ by evaluating any of the $2^k$ possible functions of the form:
\begin{align}
    f_{\text{interp}} = (\Dec{1} \circ ... \circ \Dec{k})
\end{align}
where the parameters of each \FuncName{} $\Dec{i}$ are either the parameters of $\Dec{I}$ or $\Dec{J}$. Figures~\ref{fig:interp}~and~\ref{fig:interp-comparison} show that interpolation with function composition provides interesting, meaningful outputs in experiments with $k=2$ and $k=4$.

%%%%%%%%%%%%%%%%%%%%%%%%%%%%%%%%%%%%%%%%%%%%%%%%%%%%%%%%%%%%%%%%%%%%%%%%%%%%%%%%%%%%%%%%
%%%%%%%%%%%%%%%%%%%%%%%%%%%%%%%%%%%%%%%%%%%%%%%%%%%%%%%%%%%%%%%%%%%%%%%%%%%%%%%%%%%%%%%%
\section{Experimental Evaluations}

\begin{table}
\centering
\caption{Comparing various reconstruction architectures. Reported Chamfer distance values are multiplied by 100 for readability and include standard error in parentheses. HOF-1 and HOF-3 are HOF variants with 1 and 3 hidden layers, respectively.}
\begin{tabular}{lccccc} \toprule
Method & CD(P,T) & CD(T,P) & Average CD & Parameters & Layers  \\ \midrule
3D-R2N2 & 2.588\phantom{ (0.0050)} & 3.342\phantom{ (0.0050)} & 2.965\phantom{ (0.0050)} & 2,000,000+ & 12 \\
PSG & 1.982\phantom{ (0.0050)} & 2.146\phantom{ (0.0050)} & 2.064\phantom{ (0.0050)} & --- & --- \\
EPCG & 1.846\phantom{ (0.0050)} & 1.701\phantom{ (0.0050)} & 1.774\phantom{ (0.0050)} & 60,000,000+ & 8 \\
FoldingNet & 1.563 (0.0064) & 1.534 (0.0053) & 1.549 (0.0117) & 1,056,775 & 6 \\
DeepSDF & 1.521 (0.0029) & \textbf{0.962} (0.0018) & \textbf{1.242} (0.0047) & 1,578,241 & 8 \\
HOF-1 & 1.517 (0.0040) & 1.021 (0.0028) & 1.269 (0.0068) & \phantom{1,50}\textbf{7,171} & \textbf{2} \\
HOF-3 & \textbf{1.480} (0.0028) & 1.014 (0.0020) & \textbf{1.247} (0.0048) & \phantom{1,5}34,052 & 4 \\
\bottomrule
\end{tabular}
\label{tab:mainresults}
\end{table}

%%%%%%%%%%%%%%%%%%%%%%%%%%%%%%%%%%%%%%%%%%%%%%%%%%%%%%%%%%%%%%%%%%%%%%%%%%%%%%%%%%%%%%%%
%%%%%%%%%%%%%%%%%%%%%%%%%%%%%%%%%%%%%%%%%%%%%%%%%%%%%%%%%%%%%%%%%%%%%%%%%%%%%%%%%%%%%%%%
We conduct various empirical studies in order to justify two key claims. In Sections~\ref{reconstruction_data}~and~\ref{sec:runtime}, we compare with other contextual mapping architectures to demonstrate that HOF provides equal or better performance with a significant reduction in parameters and compute time. In Section~\ref{sec:composition} we demonstrate that extending contextual mapping approaches such as HOF with multiple compositions of the mapping function provides a simple and effective approach to object interpolation. Further experimentation, including ablation studies and a simulated robot navigation scenario, can be found in~\ref{sec:additional-experiments}.

\subsection{Evaluating Reconstruction Quality on ShapeNet}
\label{reconstruction_data}
We test \method{}'s ability to reconstruct a 3D point cloud of an object given a single RGB image, comparing with other architectures for 3D reconstruction. We conduct two experiments:

\begin{enumerate}
    \item We compare HOF with LVC architectures to show that HOF is a more parameter-efficient approach to contextual mapping than existing fixed-decoder architectures.
    \item We compare HOF to a broader set of state of the art methods on a larger subset of the ShapeNet dataset, demonstrating that it matches or surpasses existing state of the art methods in terms of reconstruction quality. 
\end{enumerate}

In the first experiment, we compare HOF with other LVC architectures on 13 of the largest classes of ShapeNet~\citep{yan2016perspective}, using the asymmetric Chamfer distance metrics (Equation~\ref{eq:chamfer}) as reported in \cite{lin2018learning}. In the second experiment, we compare HOF with other methods for 3D reconstruction on a broader selection of 55 classes the ShapeNet dataset, as in~\cite{tatarchenko2019single}. In line with the recommendations in~\cite{tatarchenko2019single}, we report F1 scores for this evaluation.

\subsubsection{LVC Architecture Comparison}

For this experiment, we compare HOF with LVC decoder architectures proposed in the literature, specifically those used in DeepSDF~\citep{park2019deepsdf} and FoldingNet~\citep{yang2018foldingnet}, as well as several other baselines. Each architecture maps points from $\mathbb{R}^c$ to $\mathbb{R}^3$ in order to enable a direct comparison. The dataset contains 31773 ground truth point cloud models for training/validation and 7926 for testing. For each point cloud, there are 24 RGB renderings of the object from a fixed set of 24 camera positions. For both training and testing, each point cloud is shifted so that its bounding box center is at the origin in line with~\cite{fan2017point}. At test time, there is \textit{no post-processing performed on the predicted point cloud}. The architectures we compare in this experiment are:

\begin{enumerate}
    \item \method{}-1: 1 hidden layer containing 1024 hidden neurons
    \item \method-3: 3 hidden layers containing 128 hidden neurons
    \item DeepSDF as described in \cite{park2019deepsdf}, with 8 hidden layers containing 512 neurons each
    \item FoldingNet as described in \cite{yang2018foldingnet}, with 2 successive `folds', each with a 3-layer MLP with 512 hidden neurons
    \item EPCG architecture as reported in \cite{lin2018learning}
    \item Point Set Generation network~\citep{fan2017point} as reported in \cite{lin2018learning}
    \item 3D-R2N2~\citep{choy2016r2n2} as reported in \cite{lin2018learning}
\end{enumerate}

Results are reported in Table~\ref{tab:mainresults}. Chamfer Distance scores are scaled by 100 as in line with~\cite{lin2018learning}. We find that \method{} performs significantly better than that direct decoding baseline of \cite{lin2018learning} and performs on par with other contextual mapping approaches with $30\times$ fewer parameters. In order to provide a fair comparison with the baseline method, we ensure that ground truth objects are scaled identically to those in \cite{lin2018learning}. We report both `forward' Chamfer distance CD(Pred, Target) and `backward' Chamfer distance CD(Target, Pred), again in line with the convention established by \cite{lin2018learning}. Table~\ref{tab:results} contains a class-wise breakdown. Qualitative comparisons of the outputs of HOF with state-of-the-art architectures are shown in Figure~\ref{fig:recon-compare}.

\subsubsection{ShapeNet Breadth Comparison}

\cite{tatarchenko2019single} question the common practice in single-view 3D reconstruction of evaluating only on the largest classes of ShapeNet. The authors demonstrate that reconstruction
\begin{wraptable}{r}{5.5cm}
\centering
\caption{F1/class-weighted F1 score comparison of HOF with methods reported in~\cite{tatarchenko2019single}. Higher is better.}
\begin{tabular}{ccc} \\ \toprule
\textbf{Method} & \textbf{F1} & \textbf{cw-F1}  \\ \midrule
\textit{Oracle NN} & \textit{0.290} & \textit{0.321} \\ \midrule
AtlasNet & 0.252 & 0.287 \\ \midrule
OGN & 0.217 & 0.230 \\ \midrule
Matryoshka & 0.264 & 0.297 \\ \midrule
Retrieval & 0.237 & 0.260 \\ \midrule
HOF (Ours) & \textbf{0.291} & \textbf{0.310} \\ \bottomrule
\end{tabular}
\label{tab:what3d-results}
% \vspace{-1cm}
\end{wraptable}
methods do not exhibit performance correlated with the size of object classes, and thus evaluating on smaller ShapeNet classes is justified. We use the dataset provided by~\cite{tatarchenko2019single}, which includes 55 classes from the ShapeNet dataset. The authors also suggest using the F1 score metric, defined as the harmonic mean between precision and recall~\citep{tatarchenko2019single}.

We include comparisons with AtlasNet~\citep{groueix2018papier}, Octree Generating Networks~\citep{tatar2017octree}, Matryoshka Networks~\citep{richter2018matryoshka}, and retrieval baselines as reported by~\cite{tatarchenko2019single}. We find that HOF performs competitively with all of these state-of-the-art methods, even surpassing them on many classes. We show summary statistics in Table~\ref{tab:what3d-results}. The F1 column contains the average F1 score for each method, uniformly averaging over all classes regardless of how class imbalances in the testing set. The cw-F1 score column averages over class F1 scores weighted by the fraction of the dataset comprised by that class; that is, classes that are over-represented in the testing set are correspondingly over-represented in the cw-F1 score. On the mean F1 metric, HOF outperforms all other methods, including the `Oracle Nearest-Neighbor' approach described by~\cite{tatarchenko2019single}. The Oracle Nearest Neighbor uses the closest object in the training set as its prediction for each test sample. A complete class-wise performance breakdown is in the Appendix in Table~\ref{tab:fscore}. We find that HOF outperforms all 5 comparison methods (including the Oracle) in 23 of the 55 classes. Excluding the oracle, HOF shows the best performance in 28 of the 55 classes.

\subsection{Runtime Performance Comparison}
\label{sec:runtime}
\begin{figure}
    \centering
    \includegraphics[width=0.49\textwidth]{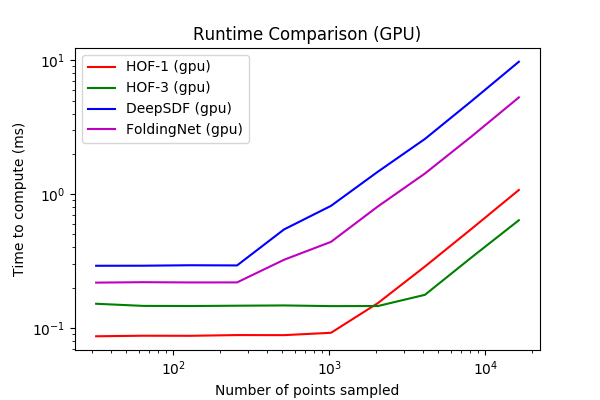}
    \includegraphics[width=0.49\textwidth]{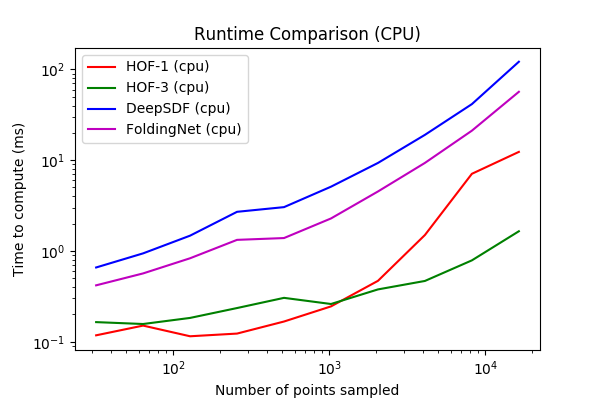}
    \caption{Runtime analysis comparing \method{} with DeepSDF and FoldingNet architectures. \method-1 and \method-3 are \method{} with 1 and 3 hidden layers, respectively. Computing environment details are given in Section~\ref{sec:implementation-details}.}
    \label{fig:performance_multiview}
\end{figure}

We compare HOF with the decoder architectures proposed in \cite{park2019deepsdf} and \cite{yang2018foldingnet} in terms of inference speed. Figure~\ref{fig:performance_multiview} shows the results of this experiment, comparing how long it takes for each network to map a set of $N$ samples from the canonical space $\set{X}$ into the object reconstruction. We ignore the processing time for estimating the latent state $\vec{z}$ for DeepSDF/FoldingNet and the function parameters $\theta$ for \method; we use the same convolutional neural network architecture with a modified output layer for both. We find that even for medium-resolution reconstructions (N > 1000), the GPU running times for the DeepSDF/FoldingNet architectures and \method{} begin to diverge. This difference is even more noticeable in the CPU running time comparison (an almost $100\times$ difference). This performance improvement may be significant for embedded systems that need to efficiently store and reconstruct 3D objects in real time; our representation is small in size, straightforward to sample uniformly (unlike a CAD model), and fast to evaluate.

% A comparison of the reconstruction quality of different \method{} architectures (and different order \FuncName s) can be found in Supplemental Section~\ref{sec:ablations}. We find that \method{} in a variety of different formulations achieves comparable accuracy to the DeepSDF architecture with far fewer parameters and much shorter running time.

% \subsection{Multi-view Reconstruction}
% \label{sec:multiview}
% In order to further motivate the composition property of HOF and to demonstrate its usefulness, we show that composing reconstruction functions estimated from different viewpoints of the same object is an effective method for aggregating the information in each viewpoint into a single reconstruction. Figure~\ref{fig:performance_multiview} (Right) demonstrates how composing HOF reconstruction functions from different viewpoints provides improved object reconstructions, surpassing the reconstruction quality from a single view.

\subsection{Object Interpolation}
\label{sec:composition}
\begin{figure}[!htb]
  \centering
  \includegraphics[width=0.49\textwidth]{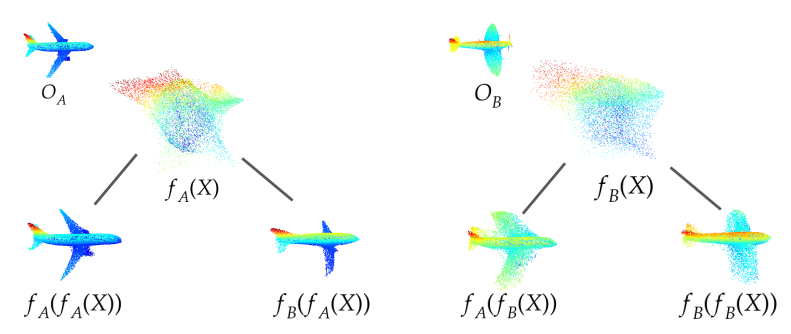} \vline 
  \includegraphics[width=0.49\textwidth]{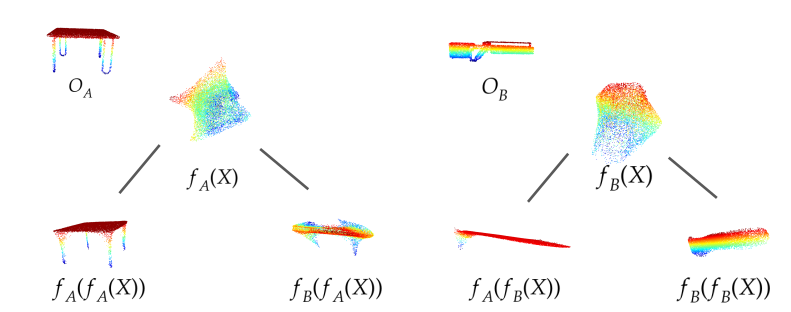} \\[4mm]
  \includegraphics[width=0.9\textwidth]{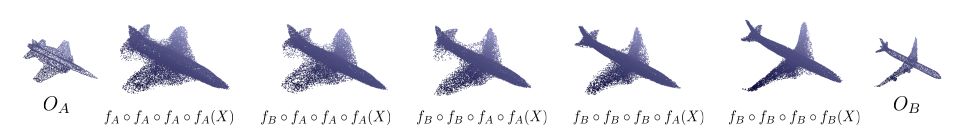}
  \caption{\textbf{Top Left.} An example of inter-class interpolation between two objects by function composition. We show the ground truth objects $\Obj_A$ and $\Obj_B$, a single evaluation of their respective decoding functions (giving $f_A(X)$ and $f_B(X)$), as well as the possible permutations of compositions, which makes up the leaf nodes in each tree. In $f_B(f_A(X))$, we see the wings straighten but remain narrow. In $f_A(f_B(X))$, we observe the wings broaden, but they remain angled. \textbf{Top Right.} An example of inter-class interpolation, mixing a table and a rifle. We observe what might be interpreted as a gun with legs in $f_B(f_A(X))$ and a table with a single coherent stock in $f_A(f_B(X))$. \\ \textbf{Bottom.} An example of intra-class interpolation between two objects with $k=4$.}
\label{fig:interp}
\end{figure}

To demonstrate that our functional representation yields an expressive latent object space, we show that the composition of these functions produces interesting, new objects.
%Figure~\ref{fig:interp3} demonstrates a smooth gradient of compositions of the reconstruction functions for two airplanes. In addition, 
The top of Figure~\ref{fig:interp} shows in detail the composition procedure. If we have estimated 2-\FuncName s for two objects $\Obj_A$ and $\Obj_B$, we demonstrate that $\Dec{A}(\Dec{B}(\set{X}))$ and $\Dec{B}(\Dec{A}(\set{X}))$ both provide interesting mixtures of the two objects and mix the features of the objects in different ways; the functions are not commutative. This approach is conceptually distinct from other object interpolation methods, which decode the interpolation of two different latent vectors. In our formulation, we visualize the outputs of an encoder that has been trained to output 2-\FuncName s in $\R^3$. In addition, the bottom of Figure~\ref{fig:interp} demonstrates a smooth gradient of compositions of the reconstruction functions for two airplanes, when a higher order of mappings ($k = 4$) is used.

% \begin{figure}[!htb]
%   \centering
%   \includegraphics[width=0.9\textwidth]{figures/interpolations/k4mapping.png}
%   \caption{An example of intra-class interpolation between two objects with $k=4$.}
% \label{fig:interp3}
% \end{figure}

To further convey the expressiveness of the composition-based object interpolation, we compare it against a method that performs interpolation in the network parameter space.
This latter approach resembles a common way of performing object interpolation in LVC methods: Generate latent codewords from each image, and synthesize new objects by feeding the interpolated latent vectors into the decoder. As a proxy for the latent vector interpolation used in LVC methods, we generate new objects as follows. After outputting the network parameters $\theta_A$ and $\theta_B$ for the objects $\Obj_A$ and $\Obj_B$, we use the interpolated parameters $\theta' = (\theta_A + \theta_B)/2$ to represent the mapping function.
In Figure~\ref{fig:interp-comparison}, we show that our composition-based interpolation is more capable of generating coherent new objects whose geometric features inherited from the source objects are preserved better.

% \begin{figure}%[!htb]
%   \centering
%   \includegraphics[width=0.49\textwidth]{figures/interpolations/interp_composition_vs_averaging1.png} {\color{gray}\vline}
%   \includegraphics[width=0.49\textwidth]{figures/interpolations/interp_composition_vs_averaging2.png}\\
%   {\color{gray}\hrule} %\hspace{0.1cm} \hline \vspace{0.1cm}
%   \includegraphics[width=0.49\textwidth]{figures/interpolations/interp_composition_vs_averaging3.png} {\color{gray}\vline} 
%   \includegraphics[width=0.49\textwidth]{figures/interpolations/interp_composition_vs_averaging6.png} 
%   \caption {Inter-class interpolation between two objects using function composition and parameter interpolation. We see that the composition-based interpolation preserves some geometric features such as chair legs, car roof, and airplane wings. Direct interpolation of the network parameters fails to meaningfully capture features from the parent objects. We use $k=4$, and $(f_B \circ f_B \circ f_A \circ f_A) (X)$ for the interpolations with composition.}
% \label{fig:interp-comparison}
% \end{figure}

%%%%%%%%%%%%%%%%%%%%%%%%%%%%%%%%%%%%%%%%%%%%%%%%%%%%%%%%%%%%%%%%%%%%%%%%%%%%%%%%%%%%%%%%
%%%%%%%%%%%%%%%%%%%%%%%%%%%%%%%%%%%%%%%%%%%%%%%%%%%%%%%%%%%%%%%%%%%%%%%%%%%%%%%%%%%%%%%%
\section{Conclusion and Future Work}
%%%%%%%%%%%%%%%%%%%%%%%%%%%%%%%%%%%%%%%%%%%%%%%%%%%%%%%%%%%%%%%%%%%%%%%%%%%%%%%%%%%%%%%%
%%%%%%%%%%%%%%%%%%%%%%%%%%%%%%%%%%%%%%%%%%%%%%%%%%%%%%%%%%%%%%%%%%%%%%%%%%%%%%%%%%%%%%%%
% In this paper, we showed that a 3D object can be efficiently encoded as a function $\Dec{}$ which maps points in $\R^3$ to the set of points that define the object.
We presented \emph{Higher Order Function Networks} (\method{}), which generate a functional representation of an object from an RGB image. The function can be represented as a small MLP with `fast-weights', or weights that are output by an encoder network $\Enc$ with learned, fixed weights. 
\method{} demonstrates state-of-the-art reconstruction quality, as measured by Chamfer distance and F1 score with ground truth models, with far fewer decoder parameters than existing methods. Additionally, we extended contextual mapping methods to allow for interpolation between objects by composing the roots of their corresponding \FuncName{} functions. Another advantage of HOF is that points on the surface can be sampled directly, without expensive post-processing methods such as estimating level sets. 
% To achieve compositionality, we modified \method{} to output a \FuncName{} $\Dec{}$ such that $\Dec{}^n,\,n>1$, rather than $\Dec{}$, produces accurate reconstructions. This formulation enables interpolation between objects by composing decoder functions estimated from different objects.

Future work might investigate the uses of HOF as a sort of `neural function currying', which could be applied to any sort of learning problem with functions of multiple variables (for example, Q-learning). In addition, the parameter-efficiency of HOF remains a challenge, and we hope to explore versions of HOF that output only a sparse but flexible subset of the parameters of the mapping function. More research is also needed in connecting HOF with other works investigating the properties of `high-quality' neural network parameters and initializations such as HyperNetworks~\citep{ha2016hyper}, the Lottery Ticket Hypothesis~\citep{frankle2018lottery}, and model-agnostic meta learning~\citep{FinnAL17}. 

There are also many interesting applications of \method{} in domains such as robotics. A demonstrative application in motion planning
can be found in Appendix~\ref{sec:path-planning-experiment}, and \cite{engin2019higher} explore extensions of HOF for multi-view reconstruction and motion planning. Using functional representations directly for example for manipulation or navigation tasks, rather than generating intermediate 3D point clouds, is also an interesting avenue of future work. We hope that the ideas presented in this paper provides a basis for future developments of efficient 3D object representations and neural network architectures.

\newpage

\bibliographystyle{unsrtnat}
\bibliography{iclr_2020}

\newpage

%%%%%%%%%%%%%%%%%%%%%%%%%%%%%%%%%%%%%%%%%%%%%%%%%%%%%%%%%%%%%%%%%%%%%%%%%%%%%%%%%%%%%%%%
%%%%%%%%%%%%%%%%%%%%%%%%%%%%%%%%%%%%%%%%%%%%%%%%%%%%%%%%%%%%%%%%%%%%%%%%%%%%%%%%%%%%%%%%
\appendix

\section{Additional Experiments}

\subsection{Architectural Variations}
\label{sec:ablations}
We compare \method{} trained on the subset of ShapeNet used in~\cite{lin2018learning} with several architectural variations, including using Resnet18 rather than our own encoder architecture, using a fast-weights decoder architecture with 6, rather than 3, hidden layers, and using tanh rather than relu in the fast-weights decoder. Results are summarized in Table~\ref{table:hof-ablation-arch}. We find that HOF is competitive in all of the formulations, although using the tanh activation function in the decoder function instead of the relu shows a small degradation in performance. Future work might investigate more deeply what principles underlie the design of fast-weights architectures.

\begin{table}[ht]
\centering
\caption{Comparisons of class-weighted asymmetric Chamfer distances (CD(P,T) and CD(T,P)) of architectural variants of \method{}-3 (with $k=1$).}
\begin{tabular}{lcccc} \toprule
Base (HOF-3) & ResNet18 Encoder & Deep Decoder & Tanh (Decoder only) \\ \midrule
1.480 / 1.014 & 1.529 / 0.989 & 1.499 / 0.998 & 1.652 / 1.066 \\ \bottomrule
\end{tabular}
\label{table:hof-ablation-arch}
\end{table}

\subsection{The Role of the Sampling Domain}

In the experimental results reported in the main paper, we sample input points from the interior of the unit sphere uniformly at random. However, we might sample other topologies as input to the mapping function. In this experiment, we compare the performance of HOF when we sample from the interior of the 3D sphere (standard configuration), the surface of the 3D sphere, the interior of the 3D cube, and the interior of the 4D sphere.

\begin{table}[ht]
\centering
\caption{Comparisons of \method{}-3 using various different canonical objects for sampling.}
\begin{tabular}{cccc} \toprule
3D Sphere & 3D Sphere (Surface) & 4D Sphere & 3D Cube \\ \midrule
1.480 / 1.014 & 1.494 / 1.263 & 1.478 / 0.912 & 1.481 / 0.999 \\ \bottomrule
\end{tabular}
\label{table:hof-ablation-shape}
\end{table}

\subsection{Comparing Different Values of k for Composition}

Here, we compare the performance of various instances of HOF when we vary the value of $k$, or number of self-compositions. We use the HOF-1 decoder architecture (1 hidden layer with 1024 neurons) and a Resnet18 encoder architecture. This encoder architecture is different from the baseline encoder used for the results in Table 1, which explains deviations in performance from those numbers. Although performance does not vary significantly, there are some differences in reconstruction quality across values of $k$. Future work might investigate how best to take use of self-compositional or recurrent decoder architectures to maximize both computational performance as well as accuracy.

\begin{table}[ht]
\centering
\caption{Comparisons of Resnet18 \method{}-1 with different values of $k$}
\begin{tabular}{cccc} \toprule
$k=1$ & $k=2$ & $k=3$ & $k=4$ \\ \midrule
1.635 / 1.008 & 1.574 / 1.000 & 1.601 / 1.081 & 1.655 / 0.953 \\
\hline
\end{tabular}
\label{table:hof-ablation-k}
\end{table}

% We compare \method{} trained with different architectures and different orders. Results are summarized in Table~\ref{table:hof-ablation-results}. We find that we achieve state-of-the-art performance across all of the formulations. We also note that for both architectures the 2-\FuncName{} outperforms the 1-\FuncName{} in forward Chamfer distance, but the 1-\FuncName{} outperforms the 2-\FuncName{} in backwards Chamfer distance. Future work might investigate the tradeoffs involved in these formulations of the mapping $\Dec{}$.

% \begin{table}[ht]
% \centering
% \caption{Comparisons of architectures and \FuncName{} orders. We compare \method{}-3, 1-\FuncName{}, \method{}-3, 2-\FuncName{}, \method{}-1, 1-\FuncName{}, and \method{}-1, 2-\FuncName{}.}
% \begin{tabular}{|c|c|c|c|c|c|}
% \hline
% \textbf{Method} & \method{}-3 1-\FuncName{} & \method{}-3 2-\FuncName{} & \method{}-1  1-\FuncName{} & \method{}-1  2-\FuncName{}  \\
% \hline \hline
% Avg CD & 1.534 / 1.046 & 1.498 / 1.078 & 1.709 / 0.993 &  1.582 / 1.089  \\
% \hline
% \end{tabular}
% \label{table:hof-ablation-results}
% \end{table}
% We compare the performance of several variants of \method{} on the reconstruction task. First, in Table~\ref{table:ablation-results}, we compare different training regimes for \method{}: canonical frame training (as reported in Table~\ref{tab:results}), camera frame training, and projection-regularized training.

\subsection{Projection Regularization}

\begin{figure}[!htb]
  \centering
  \includegraphics[width=0.9\textwidth]{figures/plane_slices/nonreg2.png}
  \includegraphics[width=0.9\textwidth]{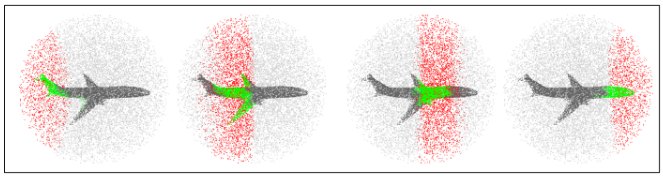}
  \caption{Slices of the sphere where the input points are sampled from and their projections in the predicted point set. Slices of the input sphere are colored red; the locations where this subset of the inputs are mapped are green. \textbf{Above}: minimizing the Chamfer distance only. \textbf{Below}: Minimizing Chamfer distance with regularization (Equation~\ref{eq:reg}). In both cases, the mapping is smooth, but only with regularization is the mapping close to the intuitive projection mapping.}
\label{fig:breadslices}
\end{figure}

Intuitively, we might expect that $\Dec{}(\set{X})$ would approximate the Euclidean projection function; e.g. $\Dec{}(\set{X}) \approx \text{Proj}_\set{Y}(X)$. However, qualitatively, we find that this is not the case. Our \FuncName{} $\Dec{}$ learns a less interpretable mapping from the canonical set $\set{X}$ to the object $\Obj$. In order to encourage the \FuncName{} to produce a more interpretable mapping from the canonical set $\set{X}$ to the object $\Obj$, we can regularize the transform $\Dec{}$ to penalize the `distance traveled' of points transformed by $\Dec{}$. A regularization term with a small coefficient ($\lambda = 0.01$) is effective in encouraging this behavior. Making this change results in little deviation in performance, while providing a more coherent mapping. Figure~\ref{fig:breadslices} highlights this distinction.

This penalty for the mapping computed by $\Dec{I}$ for each point in the sample $\tilde{\set{X}}$ is given as
\begin{equation}
\label{eq:reg}
    R(\Dec{I}, \set{\tilde{X}}) = \frac{1}{\set{\tilde{X}}}\sum_{\vec{x}_i \in \set{\tilde{X}}}||\Dec{I}(\vec{x}_i) - \vec{x}_i||_2^2
\end{equation}

where $\set{\tilde{X}}$ is a sample from the canonical set $\set{X}$. We might instead directly penalize the difference between $\Dec{I}$ and the Euclidean projection over the sampled set $\set{\tilde{X}}$ as:
\begin{equation}
    R(\Dec{I}, \set{\tilde{X}}) = \frac{1}{\set{\tilde{X}}}\sum_{\vec{x}_i \in \set{\tilde{X}}}||\Dec{I}(\vec{x}_i) - \text{argmin}_{\vec{o}_i \in \Obj}||\vec{o}_i - \vec{x}_i||_2||_2^2
\end{equation}

However, we find that this regularization can be overly constraining, for example in cases where points are sampled near the boundaries of the Voronoi tesellation of the target point cloud. The formulation in Equation~\ref{eq:reg} gives the mapping greater flexibility while still giving the desired semantics.

%%%%%%%%%%%%%%%%%%%%%%%%%%%%%%%%%%%%%%%%%%%%%%%%%%%%%%%%%%%%%%%%%%%%%%%%%%%%%%%%%%%%%%%%
%%%%%%%%%%%%%%%%%%%%%%%%%%%%%%%%%%%%%%%%%%%%%%%%%%%%%%%%%%%%%%%%%%%%%%%%%%%%%%%%%%%%%%%%
\subsection{Collision-free Path Generation}
%%%%%%%%%%%%%%%%%%%%%%%%%%%%%%%%%%%%%%%%%%%%%%%%%%%%%%%%%%%%%%%%%%%%%%%%%%%%%%%%%%%%%%%%
%%%%%%%%%%%%%%%%%%%%%%%%%%%%%%%%%%%%%%%%%%%%%%%%%%%%%%%%%%%%%%%%%%%%%%%%%%%%%%%%%%%%%%%%

From Chamfer distance or F1 scores alone, it is difficult to determine if one method's reconstruction quality is meaningfully different from another's. In this section, we introduce a new benchmark to evaluate the practical implications of using 3D reconstructions for collision-free path generation. We compare the reconstructions of \method{} with \cite{lin2018learning}, which is the most competitive direct decoding method according to Table~\ref{tab:mainresults}. This experiment is intended to give an additional perspective on what a difference in average Chamfer distance to the ground truth object means. We show that given an RGB image, we can efficiently find a near-optimal path $\hat{P}$ between two points around the bounding sphere of the object without colliding with it, and without taking a path much longer than the optimal path $P^*$, where the optimal path is defined as the shortest collision-free path between two given points. A complete definition of the experiment and its implementation are given in  Section~\ref{sec:path-planning-experiment}.

\begin{figure}[!htb]
  \centering
  \includegraphics[width=0.95\textwidth]{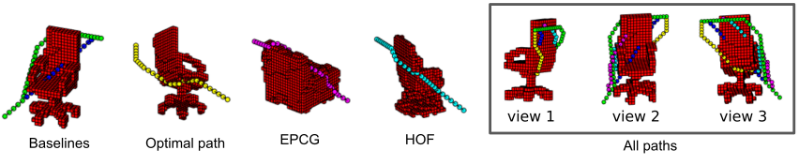}
  \caption{Collision-free path generation. The paths are color-coded as \textbf{Blue}: baseline showing $L_1$ distance, \textbf{Green}: baseline going around a bounding box around the object, \textbf{Yellow}: with GT voxels as obstacles, \textbf{Magenta}: with EPCG voxels, \textbf{Cyan}: with \method{} voxels. The rightmost figure shows all paths together viewed from three different view points. Best viewed in color.}
  \label{fig:path_cartoon}
\end{figure}

We quantify the quality of our predictions by measuring both \textit{i}) the proportion of predicted paths $\hat{P}$ that are collision-free and \textit{ii}) the average ratio of the length of a collision-free estimated path $\hat{P}$ and the corresponding optimal path $P^*$.

These two metrics conceptually mirror the backward and forward Chamfer distance metrics, respectively; a low collision rate corresponds to few missing structures in the reconstructed object (backward Chamfer, or surface coverage), while successful paths close to the optimal path length correspond to few extraneous structures in the reconstruction (forward Chamfer, or shape similarity).

% \subsection{Results}
\label{sec:path-planning-camera}
We find that \method{} provides meaningful gains over the reconstruction method recently proposed in~\cite{lin2018learning} in the context of path planning around the reconstructed model. \method{} performs significantly better both in terms of path length as well as collision rate. However, although in \cite{lin2018learning} results were reported on the reconstruction task with objects in a canonical frame, in the context of robotics, learning in a viewer-centric frame is necessary. It has been noted in \cite{wu2018learning} that generalization might be easier when learning reconstruction in a viewer-oriented frame. We test this theory by training on both objects in their canonical frame as well as in the `camera' frame. We rotate each point cloud $\set{Y}$ into its camera frame orientation using the azimuth and elevation values for each image. We rotate the point cloud about the origin, keeping the bounding box centered at (0,0,0). Trained and tested in the viewer-centric camera frame, \method{} performs even better than in the canonical frame, giving Chamfer distance scores of 1.486 / 0.979 (compared with 1.534 / 1.046 for the canonical frame). The most notably improved classes in the viewer-centric evaluation are cabinets and loudspeakers; it is intuitive that these particularly `boxy' objects might be better reconstructed in a viewer-centric frame, as their symmetric nature might make it difficult to identify their canonical frame from a single image.

Results of this comparison, as well as other ablation studies, are reported in Supplementary Tables~\ref{table:hof-ablation-arch} and \ref{table:hof-ablation-shape}.
The path quality performances of the baseline metrics, EPCG~\cite{lin2018learning} and HOF are presented in Table~\ref{table:path}.

\begin{table}[ht!]
\centering
\caption{Mean values for the collision-free path generation success rate and the optimality of the output paths. Higher values indicate better performance. Details about baseline metrics \textit{Shortest $L_1$} and \textit{Shortest Around Bounding Box} (\textit{SABB}) are listed in Supplementary Section~\ref{sec:path-planning-experiment}.}
\begin{tabular}{lllll} \toprule
Method & Shortest $L_1$ & SABB & EPCG~\citep{lin2018learning} & Ours  \\ \midrule
Success rate & 0.341 $\pm$ 0.35 & \textbf{1.0 $\pm$ 0.0} & 0.775 $\pm$ 0.19  & \textbf{0.989 $\pm$ 0.06} \\ \midrule
Optimality & \textbf{1.0 $\pm$ 0.0} & 0.960 $\pm$ 0.05  & 0.994 $\pm$ 0.02   & \textbf{0.998 $\pm$ 0.01} \\
\bottomrule
\end{tabular}
\label{table:path}
\end{table}

\subsection{Parameter Interpolation vs Function Composition}

\begin{figure}%[!htb]
  \centering
  \includegraphics[width=0.49\textwidth]{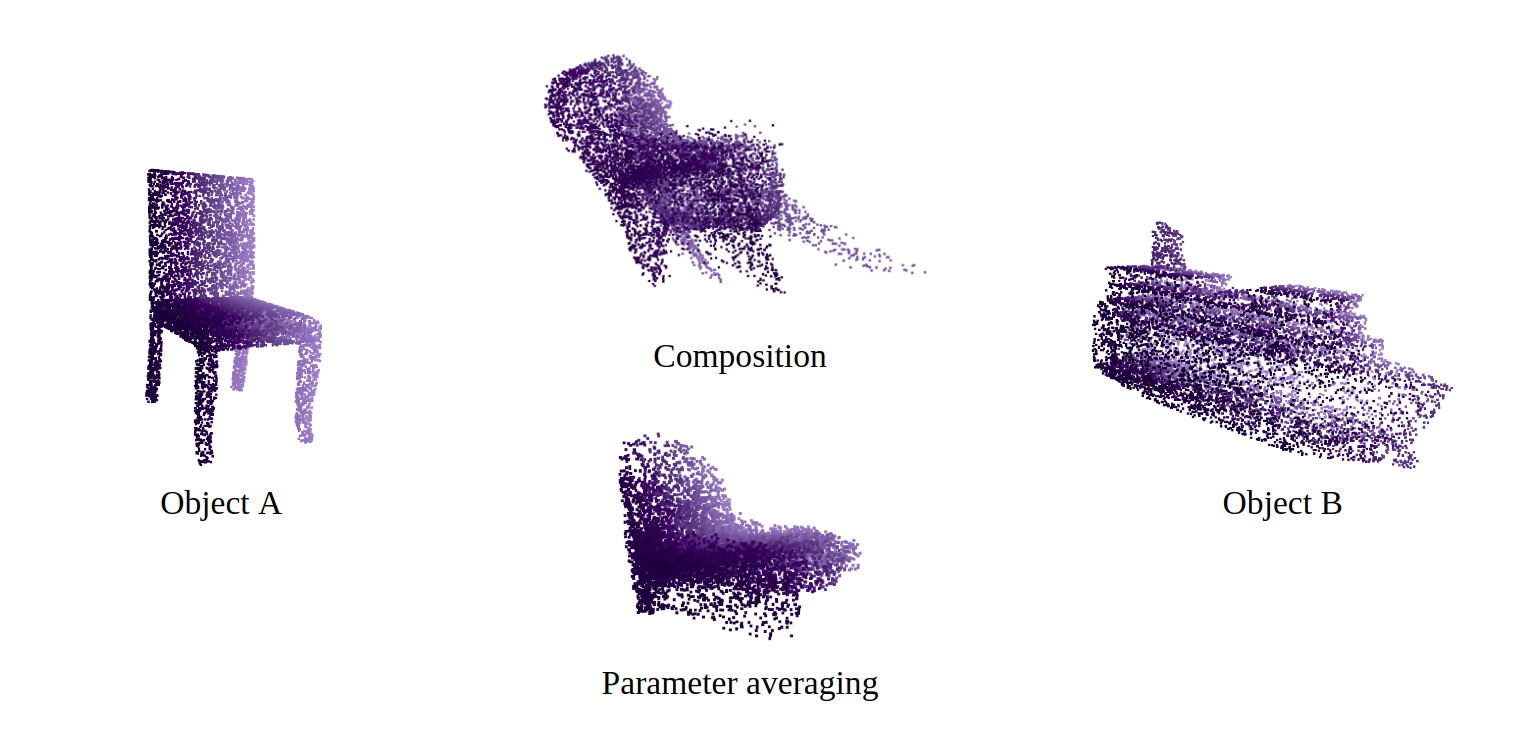} {\color{gray}\vline}
  \includegraphics[width=0.49\textwidth]{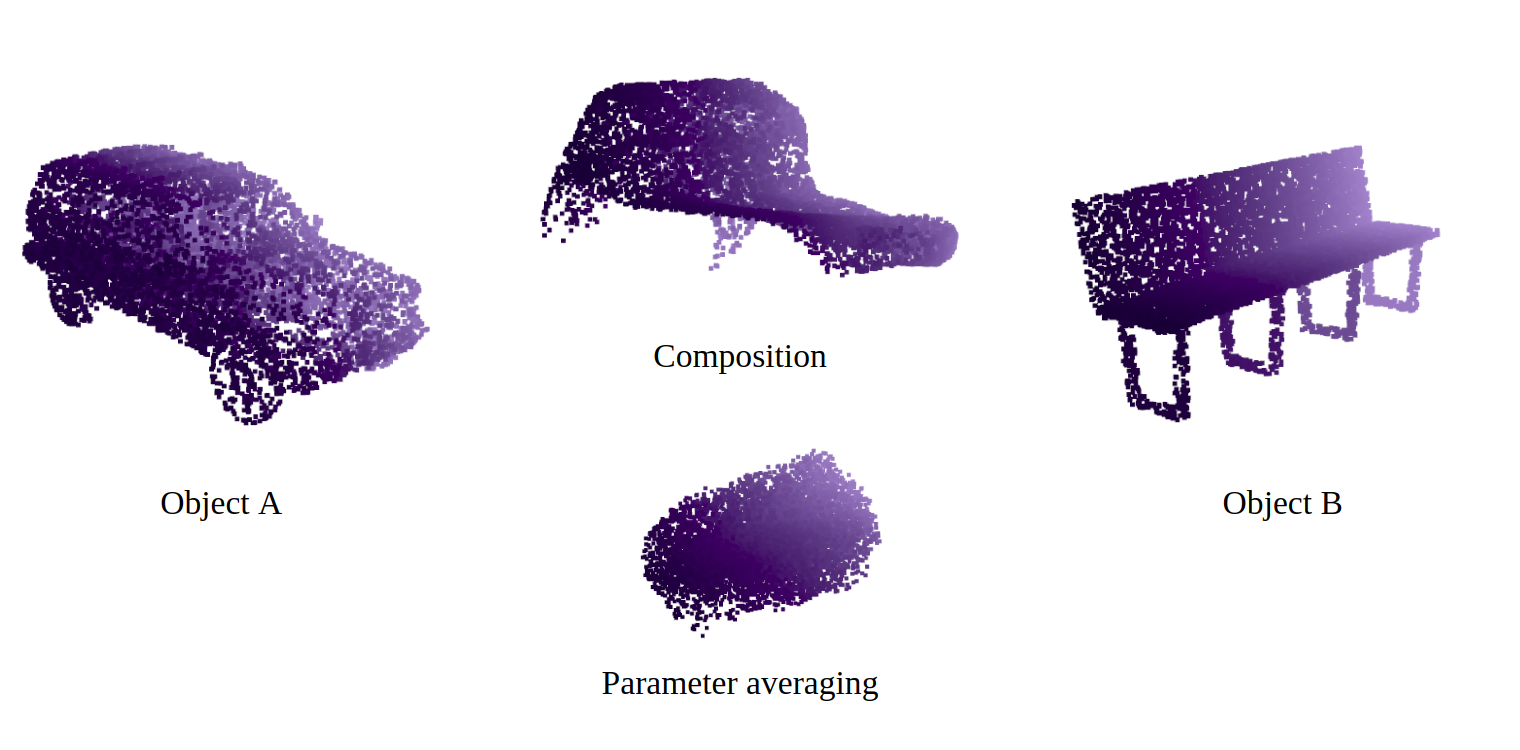}\\
  {\color{gray}\hrule} %\hspace{0.1cm} \hline \vspace{0.1cm}
  \includegraphics[width=0.49\textwidth]{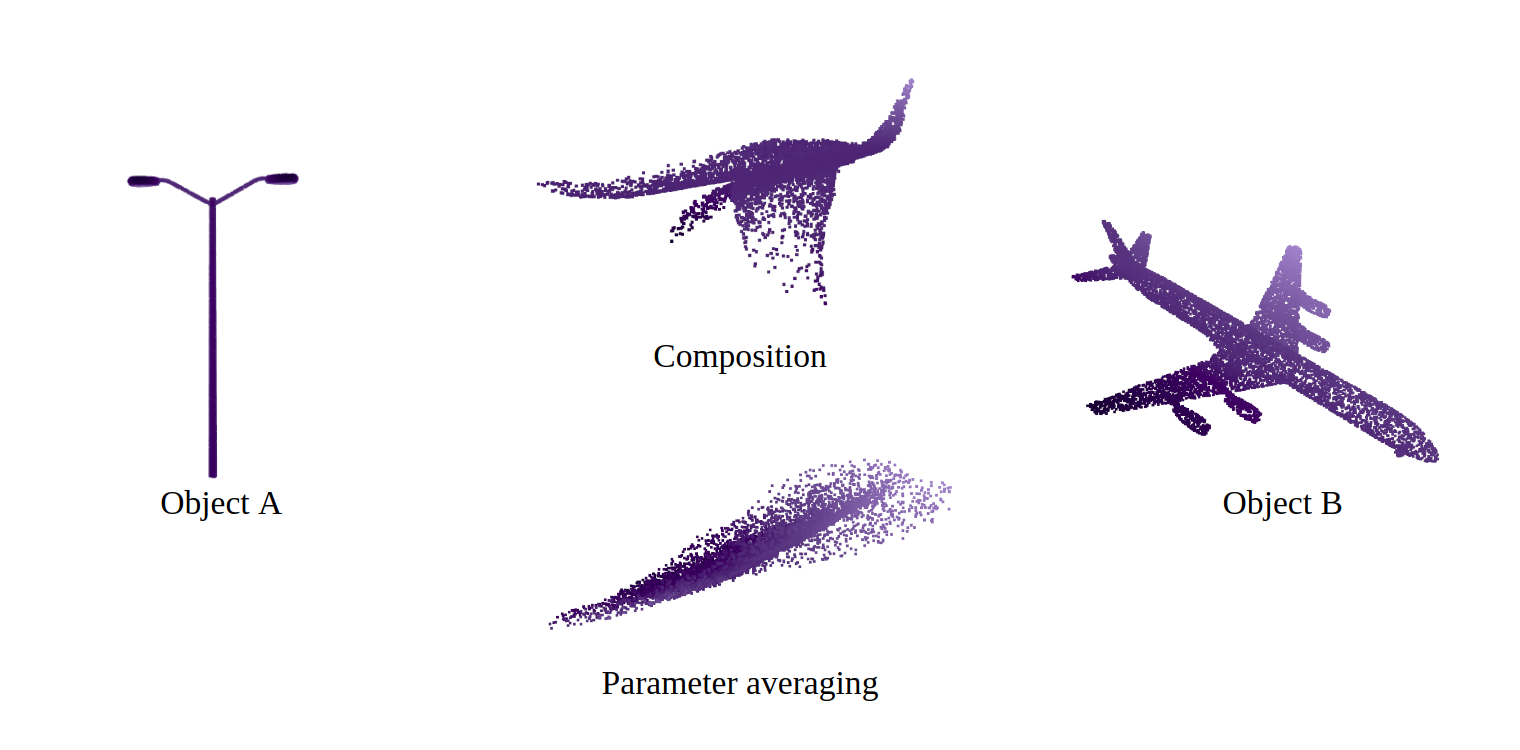} {\color{gray}\vline} 
  \includegraphics[width=0.49\textwidth]{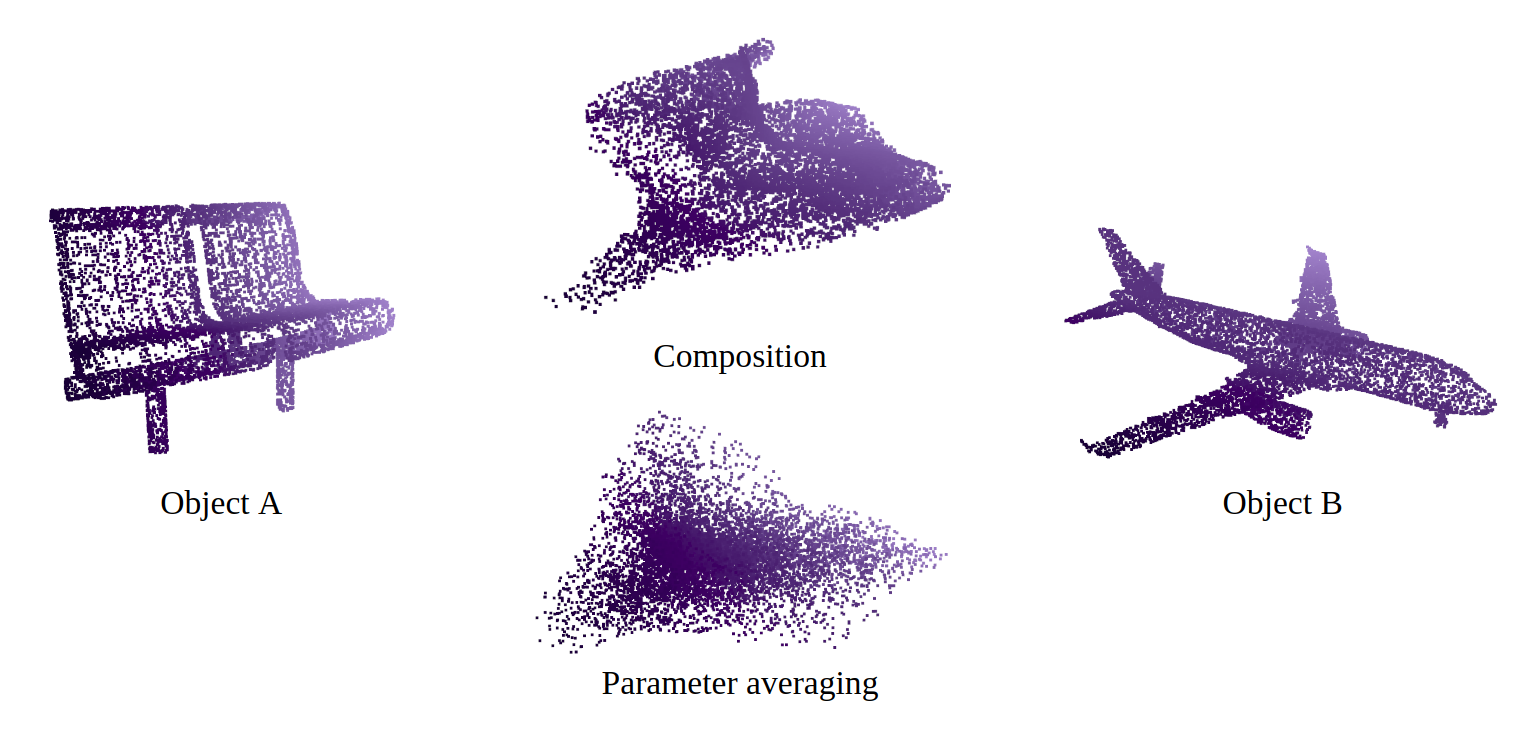} 
  \caption {Inter-class interpolation between two objects using function composition and parameter interpolation. We see that the composition-based interpolation preserves some geometric features such as chair legs, car roof, and airplane wings. Direct interpolation of the network parameters fails to meaningfully capture features from the parent objects. We use $k=4$, and $(f_B \circ f_B \circ f_A \circ f_A) (X)$ for the interpolations with composition.}
\label{fig:interp-comparison}
\end{figure}

In order to illustrate the non-trivial mappings learned by HOF for $k>1$, we compare the reconstructions acquired from interpolating naively between decoder parameters of two different objects and the reconstructions acquired by composing the two reconstruction functions. Results are shown in Figure~\ref{fig:interp-comparison}.

\subsection{Full Class Breakdown for LVC Experiment Chamfer Distance scores}
Table~\ref{tab:results} shows class-wise Chamfer Distances for HOF and the baseline methods in the LVC experiment.
\label{sec:additional-experiments}
\begin{table}[ht]
\centering
\caption{Class-weighted asymmetric Chamfer distance results for our method compared to other recent methods for 3D reconstruction from images as reported in \cite{lin2018learning}. We use the \method{}-3 architecture with $k=1$.}
\label{tab:results}
\begin{tabular}{llllll} \toprule
\textbf{Category} & 3D-R2N2 & PSG  & EPCG  & \method{} (Ours) \\ \midrule
Airplane    & 2.399 / 2.391 & 1.301 / 1.488 & 1.294 / 1.541 & \textbf{0.936} / \textbf{0.723} \\ % & 0.995 / 0.734 \\
Bench       & 2.323 / 2.603 & 1.814 / 1.983 & 1.757 / 1.487 & \textbf{1.288} / \textbf{0.914} \\ % & 1.314 / 0.945 \\
Cabinet     & \textbf{1.420} / 2.619 & 2.463 / 2.444 & 1.814 / \textbf{1.072} & 1.764 / 1.383 \\ % & 1.404 / 0.999 \\
Car         & 1.664 / 3.146 & 1.800 / 2.053 & \textbf{1.446} / 1.061 & \textbf{1.367} / \textbf{0.810} \\ % & 1.390 / 0.788 \\
Chair       & 1.854 / 3.080 & 1.887 / 2.355 & 1.886 / 2.041 & \textbf{1.670} / \textbf{1.147} \\ % & 1.718 / 1.035 \\
Display     & 2.088 / 2.953 & 1.919 / 2.334 & 2.142 / 1.440 & \textbf{1.765} / \textbf{1.130} \\ % & 1.515 / 1.031 \\
Lamp        & 5.698 / 7.331 & 2.347 / 2.212 & 2.635 / 4.459 & \textbf{2.054} / \textbf{1.325} \\ % & 1.937 / 1.080 \\
Loudspeaker & 2.487 / 4.203 & 3.215 / 2.788 & 2.371 / 1.706 & \textbf{2.126} / \textbf{1.398} \\ % & 1.790 / 1.107 \\
Rifle       & 4.193 / 2.447 & 1.316 / 1.358 & 1.289 / 1.510 & \textbf{1.066} / \textbf{0.817} \\ % & 0.867 / 0.638 \\
Sofa        & 2.306 / 3.196 & 2.592 / 2.784 & 1.917 / 1.423 & \textbf{1.666} / \textbf{1.064} \\ % & 1.683 / 1.052 \\
Table       & 2.128 / 3.134 & 1.874 / 2.229 & 1.689 / 1.620 & \textbf{1.377} / \textbf{0.979} \\ % & 1.396 / 0.935\\
Telephone   & 1.874 / 2.734 & 1.516 / 1.989 & 1.939 / 1.198 & \textbf{1.387} / \textbf{0.944} \\ % & 1.266 / 0.824 \\
Watercraft  & 3.210 / 3.614 & 1.715 / 1.877 & 1.813 / 1.550 & \textbf{1.474} / \textbf{0.967} \\ % & 1.453 / 0.880 \\
\hline
Mean        & 2.588 / 3.342 & 1.982 / 2.146 & 1.846 / 1.701 & \textbf{1.534} / \textbf{1.046} \\ % & 1.486 / 0.979 \\
\bottomrule
\end{tabular}
\end{table}

\subsection{Full Class Breakdown for Broad Experiment F1 Scores}

Table~\ref{tab:fscore} shows class-wise performance of HOF as well as each method compared in the broad ShapeNet comparison as reported by~\cite{tatarchenko2019single}.

\begin{table}[ht!]
\centering
\caption{F-score evaluation ($@1\%$) on the~\cite{tatarchenko2019single} dataset}
\label{tab:fscore}
\begin{tabular}{l c c c c c c} \toprule
%\textbf{Category} & AtlasNet & OGN~\cite{tatar2017octree}  & Matryoshka  & Retrieval~\cite{tatarchenko2019single} & Oracle NN~\cite{tatarchenko2019single} \\
\textbf{Category} & AtlasNet & OGN  & Matryoshka  & Retrieval & Oracle NN & HOF (Ours) \\ \midrule
airplane & 0.39 & 0.26 & 0.33 & 0.37 & \textbf{0.45} & 0.313 \\
ashcan & 0.18 & 0.23 & \textbf{0.26} & 0.21 & 0.24 & 0.239 \\
bag & 0.16 & 0.14 & 0.18 & 0.13 & 0.15 & \textbf{0.192} \\
basket & 0.19 & 0.16 & 0.21 & 0.15 & 0.15 & \textbf{0.217} \\
bathtub & 0.25 & 0.13 & 0.26 & 0.22 & \textbf{0.26} & 0.256 \\
bed & 0.19 & 0.12 & 0.18 & 0.15 & 0.17 & \textbf{0.195} \\
bench & 0.34 & 0.09 & 0.32 & 0.3 & 0.34 & \textbf{0.359} \\
birdhouse & 0.17 & 0.13 & \textbf{0.18} & 0.15 & 0.15 & 0.175 \\
bookshelf & 0.24 & 0.18 & 0.25 & 0.2 & 0.2 & \textbf{0.274} \\
bottle & 0.34 & 0.54 & 0.45 & 0.46 & \textbf{0.55} & 0.497 \\
bowl & 0.22 & 0.18 & 0.24 & 0.2 & 0.25 & \textbf{0.267} \\
bus & 0.35 & 0.38 & 0.41 & 0.36 & \textbf{0.44} & 0.365 \\
cabinet & 0.25 & 0.29 & \textbf{0.33} & 0.23 & 0.27 & 0.311 \\
camera & 0.13 & 0.08 & 0.12 & 0.11 & 0.12 & \textbf{0.153} \\
can & 0.23 & \textbf{0.46} & 0.44 & 0.36 & 0.44 & 0.416 \\
cap & 0.18 & 0.02 & 0.15 & 0.19 & \textbf{0.25} & 0.174 \\
car & 0.3 & 0.37 & 0.38 & 0.33 & \textbf{0.39} & 0.323 \\
cellular & 0.34 & 0.45 & 0.47 & 0.41 & \textbf{0.5} & 0.415 \\
chair & 0.25 & 0.15 & \textbf{0.27} & 0.2 & 0.23 & 0.259 \\
clock & 0.24 & 0.21 & 0.25 & 0.22 & \textbf{0.27} & 0.264 \\
dishwasher & 0.2 & 0.29 & 0.31 & 0.22 & 0.26 & \textbf{0.393} \\
display & 0.22 & 0.15 & 0.23 & 0.19 & 0.24 & \textbf{0.26} \\
earphone & 0.14 & 0.07 & 0.11 & 0.11 & 0.13 & \textbf{0.23} \\
faucet & \textbf{0.19} & 0.06 & 0.13 & 0.14 & 0.2 & 0.167 \\
file & 0.22 & 0.33 & \textbf{0.36} & 0.24 & 0.25 & 0.21 \\
guitar & 0.45 & 0.35 & 0.36 & 0.41 & \textbf{0.58} & 0.329 \\
helmet & 0.1 & 0.06 & 0.09 & 0.08 & 0.12 & \textbf{0.515} \\
jar & 0.21 & 0.22 & \textbf{0.25} & 0.19 & 0.22 & 0.114 \\
keyboard & 0.36 & 0.25 & 0.37 & 0.35 & \textbf{0.49} & 0.263 \\
knife & 0.46 & 0.26 & 0.21 & 0.37 & \textbf{0.54} & 0.504 \\
lamp & 0.26 & 0.13 & 0.2 & 0.21 & 0.27 & \textbf{0.309} \\
laptop & 0.29 & 0.21 & \textbf{0.33} & 0.26 & \textbf{0.33} & 0.284 \\
loudspeaker & 0.2 & 0.26 & \textbf{0.27} & 0.19 & 0.23 & 0.252 \\
mailbox & 0.21 & 0.2 & 0.23 & 0.2 & 0.19 & \textbf{0.292} \\
microphone & 0.23 & 0.22 & 0.19 & 0.18 & 0.21 & \textbf{0.315} \\
microwave & 0.23 & \textbf{0.36} & 0.35 & 0.22 & 0.25 & 0.279 \\
motorcycle & 0.27 & 0.12 & 0.22 & 0.24 & \textbf{0.28} & 0.27 \\
mug & 0.13 & 0.11 & 0.15 & 0.11 & \textbf{0.17} & 0.128 \\
piano & 0.17 & 0.11 & 0.16 & 0.14 & 0.17 & \textbf{0.19} \\
pillow & 0.19 & 0.14 & 0.17 & 0.18 & \textbf{0.3} & 0.227 \\
pistol & 0.29 & 0.22 & 0.23 & 0.25 & 0.3 & \textbf{0.307} \\
pot & 0.19 & 0.15 & 0.19 & 0.14 & 0.16 & \textbf{0.2} \\
printer & 0.13 & 0.11 & 0.13 & 0.11 & 0.14 & \textbf{0.17} \\
remote & 0.3 & 0.33 & 0.31 & 0.31 & 0.37 & \textbf{0.395} \\
rifle & 0.43 & 0.28 & 0.3 & 0.36 & \textbf{0.48} & 0.47 \\
rocket & 0.34 & 0.2 & 0.23 & 0.26 & 0.32 & \textbf{0.369} \\
skateboard & 0.39 & 0.11 & \textbf{0.39} & 0.35 & 0.47 & 0.384 \\
sofa & 0.24 & 0.23 & \textbf{0.27} & 0.21 & \textbf{0.27} & 0.262 \\
stove & 0.2 & 0.19 & 0.24 & 0.18 & 0.19 & \textbf{0.247} \\
table & 0.31 & 0.24 & \textbf{0.34} & 0.26 & \textbf{0.34} & 0.325 \\
telephone & 0.33 & 0.42 & \textbf{0.45} & 0.4 & 0.5 & 0.408 \\
tower & 0.24 & 0.2 & 0.25 & 0.25 & 0.25 & \textbf{0.336} \\
train & 0.34 & 0.29 & 0.3 & 0.32 & \textbf{0.38} & 0.375 \\
vessel & 0.28 & 0.19 & 0.22 & 0.23 & 0.29 & \textbf{0.299} \\
washer & 0.2 & \textbf{0.31} & \textbf{0.31} & 0.21 & 0.25 & 0.268 \\

\bottomrule
\end{tabular}
\end{table}

\section{Training/Testing Dataset and Implementation Details}
In the reconstruction experiment, Chamfer Distance scores are scaled up by 100 as in \cite{lin2018learning} for easier comparison. For the numbers reported in Table~\ref{tab:results}, we use the best performance of 3d-r2n2 (5 views as reported in \cite{lin2018learning}). In comparing with methods like FoldingNet \cite{yang2018foldingnet} and DeepSDF \cite{park2019deepsdf}, we focus on efficiency of representation rather than reconstruction quality. The performance comparison in Figure~\ref{fig:performance_multiview} and the ablation experiment in Tables~\ref{table:hof-ablation-arch} attempt to compare these architectures in this way (FoldingNet is a slightly shallower version of the DeepSDF architecture; 6 rather than 8 fully-connected layers). 
\subsection{Dataset}
We use two different datasets for evaluation. First, in Table~\ref{tab:mainresults}, we use the ShapeNet train/validation/test splits of a subset of the ShapeNet dataset \citep{chang2015shapenet} described in \cite{yan2016perspective}. The dataset can be downloaded from \texttt{https://github.com/xcyan/nips16\_PTN}. Point clouds have 100k points. Upon closer inspection, we have found that this subset includes some inconsistent/noisy labels, including:
\begin{enumerate}
    \item Inconsistency of object interior filling (e.g. some objects are only surfaces, while some have densely sampled interiors)
    \item Objects with floating text annotations that are represented in the point cloud model
    \item Objects that are inconsistently small (scaled down by a factor of 5 or more compared to other similar objects)
\end{enumerate}
Although these types of inconsistencies are rare, they are noteworthy. We used them as-is, but future contributions might include both `cleaned' and `noisy' variants of this dataset. Learning from noisy labels is an important problem but is orthogonal to 3D reconstruction.

In our second experiment, we use a broader dataset based on ShapeNet, with train and test splits taken from~\cite{tatarchenko2019single}. The dataset can be downloaded from \texttt{https://github.com/lmb-freiburg/what3d}.

\subsection{Implementation Details}
\label{sec:implementation-details}
\subsubsection{Network Architecture and Training}
\label{sec:architecture}
For the problem of 3D reconstruction from an RGB image, which we address here, we represent $\Enc$ as a convolutional neural network based on the DenseNet architecture proposed in \cite{huang2016dense}. We call this our baseline encoder network. The baseline encoder network has 3 dense blocks (each containing 4 convolutional layers) followed by 3 fully connected layers. The schedule of feature maps is [16, 32, 64] for the dense blocks. Each fully connected layer contains 1024 neurons.

We use two variants of the \FuncName{} architecture $\Dec{}$. One, which we call \method{}-1, is an MLP with 1 hidden layer containing 1024 neurons. A second version, \method{}-3, is an MLP with 3 hidden layers, each containing 128 hidden units. Both formulations use the ReLU activation function \cite{glorot2011deep}. Because $\Enc$ and $\Dec{}$ are all differentiable almost everywhere, we can train the entire system end-to-end with backpropagation. We use the Adam Optimizer with learning rate 1e-5 and batch size 1, training for 4 epochs for all experiments (1 epoch $\approx$ 725k parameter updates). Training \method{} from scratch took roughly 36 hours.

\subsubsection{Path Planning Experiment}
\label{sec:path-planning-experiment}
We use the class `chair' from the dataset described in Section~\ref{reconstruction_data} in our experiments. The objects from this class have considerable variation and complexity in shape, thus they are useful for evaluating the quality of the generated paths.

Path planning is performed in a three dimensional grid environment. All the objects in our dataset fit inside the unit cube.
Given the predicted point cloud of an object, we first voxelize the points by constructing an occupancy map $V = n \times n \times n$ centered at the origin of the object with voxel size $2/n$. Next, we generate start and end points as follows.
We choose a unit vector $\vec{v}$ sampled uniformly at random and compute $\vec{d} = n/2 \cdot \vec{v}/||\vec{v}||_1$. We use the end points of $\vec{d}$ and $-\vec{d}$ as the start and goal locations. 
For each method, we generate the paths with the A* algorithm using the voxelization of the predicted point clouds as obstacles, and the sampled start and goal positions. The movement is rectilinear in the voxel space and the distances are measured with the $L_1$ metric. In the experiments we use an occupancy grid of size $32 \times 32 \times 32$, and sample 100 start and goal location pairs per model.

\begin{figure}[!htb]
  \centering
  \includegraphics[width=0.6\textwidth]{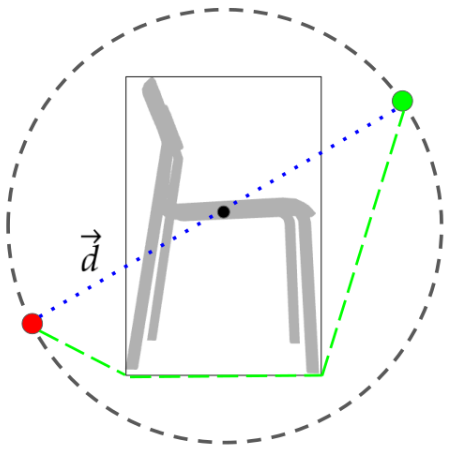}
  \caption{2D illustration of the baseline path generation methods. The dotted path in blue is produced by \textit{Shortest $L_1$} and the dashed path in green is by \textit{SABB}. In our experiments, we use the rectilinear shortest path as the output of \textit{Shortest $L_1$}.}
  \label{fig:path_2d}
\end{figure}

In addition to the paths generated using the predictions from the EPCG~\cite{lin2018learning} and \method{} methods, we present two other baselines (Figure~\ref{fig:path_2d}). The first baseline \textit{Shortest $L_1$} outputs the shortest path with the $L_1$ metric ignoring the obstacles in the scene, and the second baseline \textit{Shortest Around Bounding Box} (\textit{SABB}) takes the bounding box of the ground truth voxels as the environment to generate the path.

We present the path generation results in Table~\ref{table:path}. The baseline \textit{Shortest $L_1$} gives the optimal solution when the path is collision-free. However, since most of the shortest paths go through the object, this baseline has a poor success rate performance.
In contrast, \textit{SABB} output paths are always collision-free as the shortest path is computed using the bounding box of the true voxelization as the obstacles in the environment. The length of the paths generated by \textit{SABB} are longer compared to the rest of the methods since the produced paths are `cautious' to not collide with the object. These two baselines are the best performers for the metric they are designed for, yet they suffer from the complementary metric. Our method on the other hand achieves almost optimal results in both metrics due to the good quality reconstructions.

\subsubsection{Computing Environment}
All GPU experiments were performed on NVIDIA GTX 1080 Ti GPUs. The CPU running times were computed on one of 12 cores of an Intel 7920X processor.

% \section{Other Visualizations}
% %%%%%%%%%%%%%%%%%%%%%%%%%%%%%%%%%%%%%%%%%%%%%%%%%%%%%%%%%%%%%%%%%%%%%%%%%%%%%%%%%%%%%%%%
% %%%%%%%%%%%%%%%%%%%%%%%%%%%%%%%%%%%%%%%%%%%%%%%%%%%%%%%%%%%%%%%%%%%%%%%%%%%%%%%%%%%%%%%%

% \begin{figure}[ht]
%     \centering
%     \includegraphics[width=0.32\textwidth]{figures/quivers/quiver1.png}
%     \includegraphics[width=0.32\textwidth]{figures/quivers/quiver2.png}
%     \includegraphics[width=0.32\textwidth]{figures/quivers/quiver3.png}
%     \caption{Visualization of the \FuncName{} estimated for various objects as the scaled flow direction of each point in the input canonical space $\set{X}$. Vectors are colored according to azimuthal angle.}
% \end{figure}

\end{document}